# High-resolution rainfall-runoff modeling using graph neural network

**Zhongrun Xiang**
University of Iowa
Iowa City, IA
zhongrun-xiang@uiowa.edu

**Ibrahim Demir**
University of Iowa
Iowa City, IA
ibrahim-demir@uiowa.edu

## Abstract

Time-series modeling has shown great promise in recent studies using the latest deep learning algorithms such as LSTM (Long Short-Term Memory). These studies primarily focused on watershed-scale rainfall-runoff modeling or streamflow forecasting, but the majority of them only considered a single watershed as a unit. Although this simplification is very effective, it does not take into account spatial information, which could result in significant errors in large watersheds. Several studies investigated the use of GNN (Graph Neural Networks) for data integration by decomposing a large watershed into multiple sub-watersheds, but each sub-watershed is still treated as a whole, and the geoinformation contained within the watershed is not fully utilized. In this paper, we propose the GNRRM (Graph Neural Rainfall-Runoff Model), a novel deep learning model that makes full use of spatial information from high-resolution precipitation data, including flow direction and geographic information. When compared to baseline models, GNRRM has less over-fitting and significantly improves model performance. Our findings support the importance of hydrological data in deep learning-based rainfall-runoff modeling, and we encourage researchers to include more domain knowledge in their models.

## 1 Introduction

In recent years, deep learning models have been widely used in hydrology research [1, 2]. Because rainfall-runoff modeling is a time-series task, the LSTM (Long Short-Term Memory) advanced deep learning model has been used for hourly [3, 4], daily [5, 6], and monthly [7] modeling since 2018. When compared to other machine learning and data-driven models, studies using LSTM show a significant improvement, prompting many researchers to investigate the potential of deep learning models. These studies, on the other hand, developed lumped models by treating each watershed as a separate entity, and no geospatial data was used [8]. PRISM and NCEP/EMC Stage IV, for example, provide daily and hourly precipitation data at 4km grids. The studies using watershed scale models simply took the average precipitation of all the watershed areas, ignoring the spatial information, when modeling watersheds with hundreds of square kilometers without internal streamgage. In large watersheds, the uneven spatial distribution of precipitation may cause significant errors.

The CNN (Convolutional Neural Network) is a deep learning model that can handle data with spatial features. Since 2012 [9], CNNs have been studied and used in a variety of fields, including face recognition, medical image analysis, and object detection [10]. A convolution kernel will be applied to each node or pixel, and each node's neighbors will be calculated using the same convolution operation. However, theoretically, these models cannot work on high-resolution hydrology modeling. To address the problem, GNN is proposed as a method for capturing graph dependence via message passing between graph nodes [11]. We can convert the water flow direction map into a unidirectional directed graph and use it to build a neural network.

In this study, we proposed a novel GNN model for fully distributed rainfall-runoff simulation using high-resolution precipitation data to effectively use the spatial rainfall data and reduce the error caused by the uneven spatial distribution of precipitation.

## 2  Problem Definition

Because the original high-resolution precipitation data is in grids, we proposed our graph neural network in a fully distributed structure. Figure 1a shows an example fully distributed flow direction map of an irregularly shaped watershed. For two reasons, we can't use 2D-CNN models. To begin with, not all neighbors are physically significant. Water from precipitation at grid cells a and b, for example, will only flow to the top left of the grid cell and will not contribute to the stream in the watershed in yellow. As a result, the neighbors a and b should not be taken into account when calculating precipitation at cell c. Second, even when neighbors have the same positional relationship, they are not all equally valuable. Grid cell c, for example, is on top of cell e, and grid cell d is on top of cell f. The patterns between cells c and e are clearly different from the patterns between cells d and f, as evidenced by the water flow direction map. As a result, a water flow direction map can be converted into a graph that represents a watershed. The converted graph G(V,E) is shown in Figure 1b, with 8 nodes representing different land grid cells and the edges representing different runoff channels.

The d8 algorithms can be used to calculate the flow direction map. It is a kernel method that directs flow in one of the eight directions available [12]. We created the flow direction map in both directions in this study. Each grid cell is then treated as a node, and each flow direction is treated as an edge, resulting in a graph. In d8 algorithms, we assume that each graph edge has the same distance. We made this assumption because sequence models, such as LSTM, can only work on data with the same time step in the sequence. Hourly rainfall and effective area in each grid cell are natural attributes that can be used as node features for a watershed. We defined the problem of predicting watershed runoff at hour t using grided hourly rainfall data of the watershed from hour t-71 to t based on previous research and works.

## 3  Model Design of GNRRM

Based on the design principles of graph neural networks, we proposed a generic rainfall-runoff model using a graph neural network named Graph Neural Rainfall-Runoff Models (GNRRM). There are two key points in our design: *k*-hop with *k* value at the maximum distance, and the use of sequence model for spatial information.

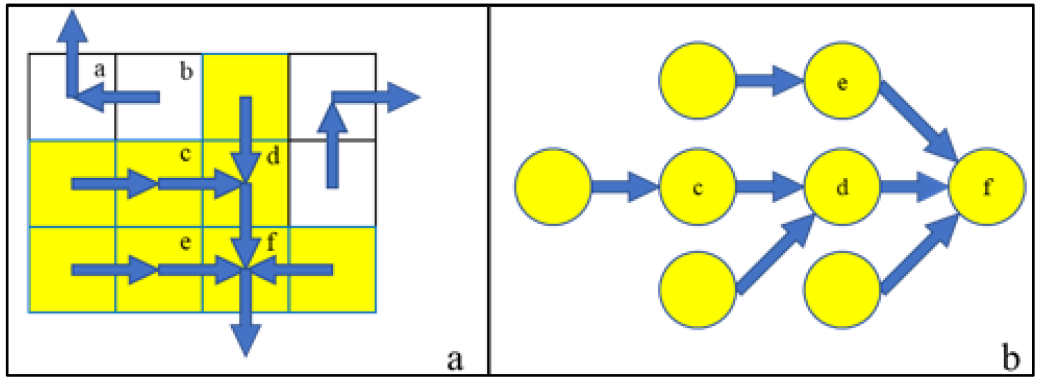

Figure 1: Example watershed with area shaded in yellow, and a streamflow outlet gauge at grid *f*. a) the original watershed in grids. b) the converted unidirectional directed graph G.

In GNN studies, the radius k neighborhood is referred to as the k-hop neighborhood. Some traditional graph research tasks, such as social networks and traffic networks, only consider low order similarity, which means they only consider neighborhoods that are one hop or less than three hops away. Because high-order neighborhoods are meaningless in many graph network tasks, this is reasonable. According to the six-degrees-of-separation theory, anyone's 6-hop friend could be anyone on the planet, which is meaningless. Because traffic congestion is primarily a short-term issue and traffic congestion over 100 kilometers is uncommon, engineering studies such as traffic predictions [13] used at most 3-hop neighborhoods in their studies. In hydrology, however, precipitation that falls on land runs off into streams, affecting downstream more or less [14]. As a result, it is critical to account for all of the rainfall in the watershed in order to make it physically meaningful. As a result, in our GNRRM, we make the k the maximum distance. Second, on different hop neighborhoods, we use the same equations and parameters. The parameters on each hop neighborhood in most GNN models, including the most popular models Graph Convolutional Networks [15] and GraphSAGE [16], are designed to be different values for a stronger characterization capability. However, in hydrology, as in many distributed hydrologic models, it is a natural way to consider each step between the neighborhoods to be the same.

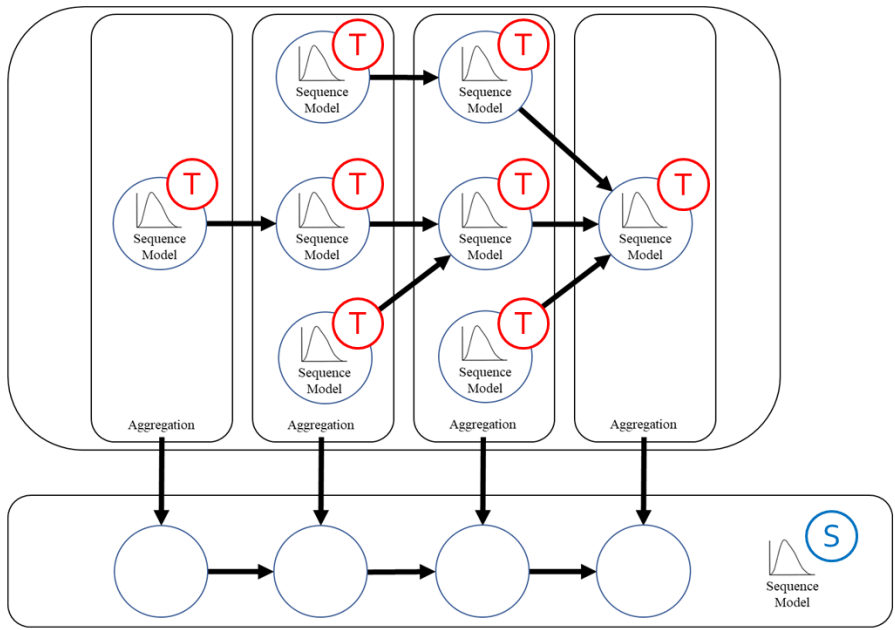

Figure 2: GNRRM model structure on the example watershed. The calculation starts from the temporal sequence model for the rainfall-runoff simulation on each land. After appropriate aggregation, the spatial sequence model is used for the spatial level modeling.

The design of our GNRRM is shown in Figure 2, this diagram is designed for the sample watershed. The calculation process is Sequence-Aggregation-Sequence. At first, we apply one sequence model on each node for the time-series data. This represents the physical processes that happened on the rainfall-runoff processes on each grid land. Then, we aggregated our data in nodes based on its distance to the outlet node using d8 flow algorithm. In detail, we create a simple hierarchical graph representation of the original graph based on the distance. And the node aggregation operations can be the sum for the area, and the weighted average for precipitation intensity. Finally, we applied the second sequence model on the hierarchical graph for spatial modeling. A sequence model such as LSTM naturally applied the same equation and calculation between nodes, which represent the water flow between two nodes is the same. In this study, we tested four basic sequence models in GNRRM. They are LSTM [17], Gated Temporal Convolution Network (GTCN) [18], and bidirectional LSTM (BiLSTM) and bidirectional GTCN (BiGTCN).

## 4  Case Study

We tested our GNRRM models on an Iowa watershed in this study. It's in the east part of Iowa, and the USGS gauge at the drainage outlet is named "USGS 05418400 North Fork Maquoketa River near Fulton, IA" and is located at Latitude 42°09'51.6", Longitude 90°43'45.6". (USGS, 2021). It is a large watershed (over 1,000 km$^2$) with a drainage area of 1,308 km$^2$. Based on the Stage IV 4-km rainfall grid, there are 104 gird cells, as shown in Figure 3. From the water year 2012 to 2018, we conducted research for seven years. The training data comes from the water year 2014-2017, while the validation data comes from the water year 2012-2013. The test is based on the 2018 water year. We only considered April and November each year due to a lack of accurate ice, snow, and spring melt snow data in high spatial and temporal resolution. We have 40,404 hourly data instances in total, which are divided into 23,137 for training, 11,491 for validation, and 5,776 for testing.

We used several simplified approaches as baseline models in addition to the proposed GNRRM. All of the features in the simple sequence model were used in the first baseline model (All). For example, in each time step of the All-LSTM model, we used a simple LSTM model with 104 features representing 104 grid cells. The only feature of the input of the sequence model in the second baseline model (Lumped) was the average rainfall intensity of the watershed each hour. The third baseline model (CNN) used a 1D convolution on the 104 grid features, with the CNN output being used as the input to sequence models. Table 2 shows the results of the GNRRM and baseline models.

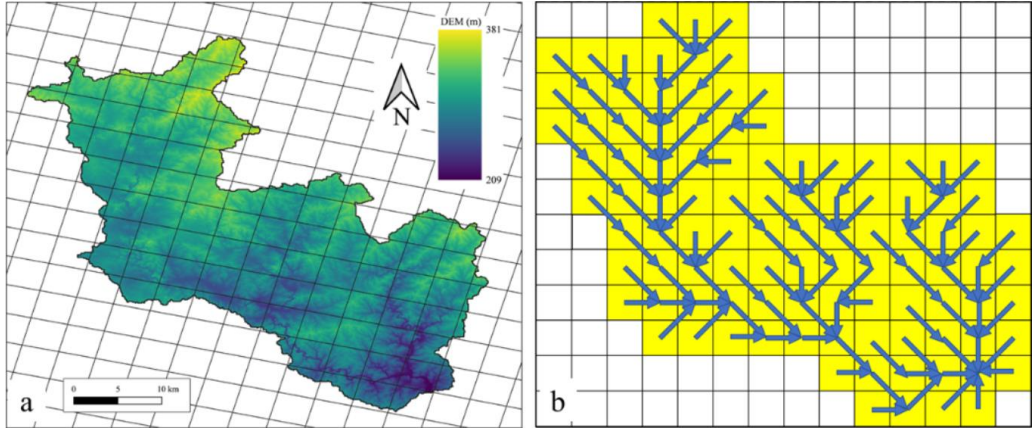

Figure 3: Study area DEM (a) and the graph representation (b) using d8 algorithm in 4-km grids.

Table 1: The dataset used in this study.

| Datasets | Data Type | Sources | Spatial Resolution | Temporal Resolution | Unit |
|---|---|---|---|---|---|
| DEM | GIS shapefile | IFC | 90-m grid | constant | m |
| Drainage area polygon | GIS shapefile | IFC | Polygon | constant | - |
| Precipitation intensity | Stage IV multi-sensor | NOAA | 4-km grid | 60-min | mm/hr |
| Streamflow rate | USGS gage measurement | USGS | Point | 15-min | m$^3$/s |

Table 2: Statistics including KGE, NSE, and RMSE on the model performance on the test dataset.

| Sequence Models | KGE | | | | NSE | | | | RMSE [m$^3$/s] | | | |
|---|---|---|---|---|---|---|---|---|---|---|---|---|
| | All | Lumped | CNN | GNRRM | All | Lumped | CNN | GNRRM | All | Lumped | CNN | GNRRM |
| LSTM | 0.510 | 0.676 | 0.657 | 0.741 | 0.175 | 0.527 | 0.543 | 0.604 | 9.853 | 7.458 | 7.333 | 6.829 |
| BiLSTM | 0.487 | 0.778 | 0.804 | 0.809 | 0.356 | 0.620 | 0.695 | 0.630 | 8.702 | 6.688 | 5.989 | 6.599 |
| GTCN | 0.553 | 0.699 | 0.736 | 0.774 | 0.561 | 0.641 | 0.632 | 0.669 | 7.184 | 6.449 | 6.582 | 6.238 |
| BiGTCN | 0.634 | 0.704 | 0.708 | **0.842** | 0.580 | 0.559 | 0.585 | **0.714** | 7.032 | 7.203 | 6.211 | **5.799** |

We find that bidirectional models outperform unidirectional models when we compare the models. We discovered that the lumped models using watershed-level averaged precipitation performed significantly better than the models using all precipitation data when we compared the input pretreatment methods. This could be explained by a correlation coefficient analysis. According to our analysis of Pearson pairwise correlation coefficients between precipitation intensity values in 104 grid cells, 96.9% of Pearson pairwise correlation coefficients are greater than 0.7, which is extremely high. The high correlation would result in a multicollinearity problem, and there is no doubt that high linear intercorrelation between the input variables could lead to incorrect regression model results. In most cases, CNN models outperform lumped models. However, the proposed GNRRM outperforms CNN and lumped models in terms of all evaluation metrics.

## 5   Conclusion

This study proposed GNRRM, a new type of network architecture for rainfall-runoff modeling based on the idea of graph neural networks, to use the geospatial information from the precipitation data. We used out-of-state time-series models like LSTM, BiLSTM, GTCN, and BiGTCN to apply baseline models. Models with designed graph architectures effectively used spatial information and avoided overfitting issues caused by spatial correlation, according to the findings. We highly recommend researchers working on the integration of physical models with current deep learning models, as the successful design of GNRRM is inspired by traditional physical rainfall-runoff models.